# Loopy Belief Propagation in Bayesian Networks : origin and possibilistic perspectives

**Amen Ajroud [1], Mohamed Nazih Omri [2], Habib Youssef [3], Salem Benferhat [4]**

[1] ISET Sousse, TUNISIA – amen.ajroud@isetso.rnu.tn
[2] IPEIM Monastir - University of Monastir, TUNISIA – nazih.omri@ipeim.rnu.tn
[3] ISITC Hammam-Sousse - University of Monastir, TUNISIA – habib.youssef@fsm.rnu.tn
[4] CRIL, Lens - University of Artois, FRANCE – benferhat@cril.univ-artois.fr

Abstract :
In this paper we present a synthesis of the work performed on two inference algorithms: the Pearl's belief propagation (BP) algorithm applied to Bayesian networks without loops (i.e. polytree) and the Loopy belief propagation (LBP) algorithm (inspired from the BP) which is applied to networks containing undirected cycles. It is known that the BP algorithm, applied to Bayesian networks with loops, gives incorrect numerical results i.e. incorrect posterior probabilities. Murphy and al. [7] find that the LBP algorithm converges on several networks and when this occurs, LBP gives a good approximation of the exact posterior probabilities. However this algorithm presents an oscillatory behaviour when it is applied to QMR (Quick Medical Reference) network [15]. This phenomenon prevents the LBP algorithm from converging towards a good approximation of posterior probabilities. We believe that the translation of the inference computation problem from the probabilistic framework to the possibilistic framework will allow performance improvement of LBP algorithm. We hope that an adaptation of this algorithm to a possibilistic causal network will show an improvement of the convergence of LBP.

## 1. Review of Bayesian Networks

Bayesian networks are powerful tools for modelling causes and effects in a wide variety of domains. They use graphs capturing causality notion between variables, and probability theory to express the causality power.

Bayesian networks are very effective for modelling situations where some information is already known and incoming data is uncertain or partially unavailable. These networks also offer consistent semantics for representing causes and effects via an intuitive graphical representation. Because of all these capabilities, Bayesian networks are regarded as systems for uncertain knowledge representation and have a large number of applications with efficient algorithms and have strong theoretical foundations [9],[10],[11] and [12].

Theoretically, a Bayesian network is a directed acyclic graph (DAG) made up of nodes and causal edges. Each node has a probability of having a certain value. Nodes are often binary, though a Bayesian network may have n-ary nodes. Parent and child nodes are defined as follows: a directed edge exists from a parent to a child. Each child node will have a conditional probability table (CPT) based on parental values. There are no directed cycles in the graph, though there may be "loops", or undirected cycles. An example network is shown in Figure 1, with parents $U_i$ sharing a child X. The node X is a child of the $U_i$'s as well as being a parent to the $Y_i$'s.

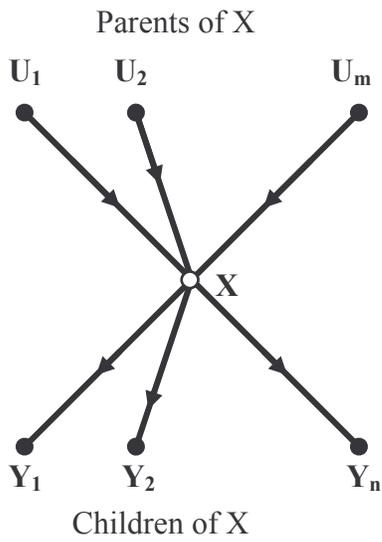

Figure 1: A Bayesian Network (polytree)

## 2. Bayesian Network inference

In bayesian network, the objective of inference is to compute *P(X|E)*, the posterior probabilities of some "query" nodes (noted by X) given some observed value of evidence nodes (noted by E, E⊄X) [13]. A simple form of it results when X is a single node, i.e., computing the posterior marginal probabilities of a single query node.

With the constitution of a large Bayesian network, the feasibility of the probabilistic inference is tested: when the network is simple, the calculation of these probabilities is not very difficult. On the other hand, when the network becomes very large, several problems emerge: the inference requires an enormous memory size and calculation becomes very complex or even, in certain cases, can not be completed.

The inference algorithms are classified in two groups [5]:

- Exact algorithms: these methods use the conditional independence contained in the networks and give, to each inference, the exact posterior probabilities. The exact probabilistic inference in general has been proven to be NP-hard by Cooper [4].

- Approximate algorithms: they are the alternatives of the exact algorithms when the networks become very complex. They estimate the posterior probabilities in various ways. Approximating probabilistic inference was also shown to be NP-hard by Dagum and Luby [8].

## 3. Pearl's algorithm

At the beginning of the 80s, Pearl published an efficient message propagation inference algorithm for polytrees [6] and [10]. This algorithm is an exact belief propagation procedure but works only for polytrees [12]. Consider the case of a general discrete node X having parents $U_1…U_m$ and children $Y_1…Y_n$, as shown in Figure 1. Evidence will be represented by E, with evidence "above" X (evidence at ancestors of X) represented as $e^+$ and evidence "below" X (evidence at descendants of X) as $e^-$.

Knowledge of evidence can flow either down the network (from parent to child) or up the network (child to parent). We use the notation $\pi(U_i)$ to represent a message from parent $U_i$ to child and $\lambda(Y_j)$ for a message from child $Y_j$ to parent (see Figure 2) [13].

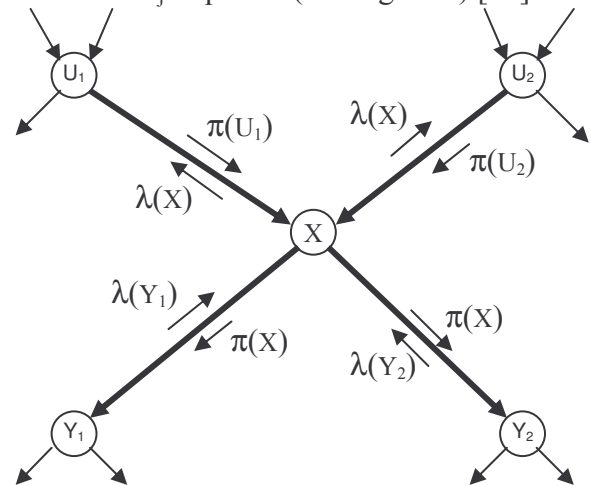

Figure 2: Messages propagation

The posterior probability (belief) on node X, given its parents and children value, can be computed as follows:

$$BEL(X) = P(X=x|e) = \alpha \lambda(X) \pi(X)$$

Where $\alpha$ is a normalizing constant, $\lambda(X) = P(e^-|x)$ and $\pi(X) = P(x|e^+)$.

To calculate $\lambda(X) = P(e^-|x)$, it is assumed that node X has received all $\lambda$ messages from its $c$ children. $\lambda_{AB}(X)$ will represent the $\lambda$ message sent from node A to node B.

$$\lambda(x) = \prod_{j=1}^{c} \lambda_{Y_j X_i}(x)$$

In the same way, in computing $\pi(X) = P(x|e^+)$, we assume that node X has received all $\pi$ messages from its $p$ parents. Similarly, $\pi_{AB}(X)$ represents the $\pi$ message sent from node A to node B. By applying the summation over the CPT of node X, we can express $\pi(X)$:

$$\pi(x) = \sum_{u_1,...,u_p} P(x|u_1,...,u_p) \prod_{j=1}^{p} \pi_{U_j X_i}(u_j)$$

Thus, we need to compute:

$$\pi_{XY_j}(x) = \alpha \pi_X(x) \prod_{k \neq j} \lambda_{Y_k X}(x)$$

and

$$\lambda_{Y_j X}(x) = \sum_{y_j} \lambda_{Y_j}(y_j) \sum_{v_1,...,v_q} p(y|v_1,...,v_q) \prod_{k=1}^{q} \pi_{V_k Y_j}(v_k)$$

To summarize, Pearl's algorithm proceeds as follows:

*A- Initialization step*
- For all nodes $V_i = e_i$ in E:
$\lambda(x_i) = 1$ wherever $x_i = e_i$; 0 otherwise
$\pi(x_i) = 1$ wherever $x_i = e_i$; 0 otherwise
- For nodes without parents:
$\pi(x_i) = p(x_i)$ - prior probabilities
- For nodes without children:
$\lambda(x_i) = 1$ uniformly (normalize at end)

*B- Iterate until no change occurs*
- (For each node X) if X has received all the $\pi$ messages from its parents, calculate $\pi(x)$
- (For each node X) if X has received all the $\lambda$ messages from its children, calculate $\lambda(x)$
- (For each node X) if $\pi(x)$ has been calculated and X received all the $\lambda$ messages from all its children (except Y), calculate $\pi_{XY}(x)$ and send it to Y.
- (For each node X) if $\lambda(x)$ has been calculated and X received all the $\pi$ messages from all parents (except U), calculate $\lambda_{XU}(x)$ and send it to U.

*C- Compute BEL(X)=$\lambda(x)\pi(x)$ and normalize*

The belief propagation algorithm has polynomial complexity in the number of nodes and converges in time proportional to the diameter of network [12]. In addition, computation in a node is proportional to its CPT size. We point out that the exact probabilistic inference in general has been proven to be NP-hard [4].

Actually, the majority of Bayesian networks are not polytrees. It is proven that Pearl's algorithm, applied to Bayesian networks with loops, gives incorrect numerical results i.e. incorrect posterior probabilities.

Pearl proposed an exact inference algorithm for multiply connected networks called "loop cutset conditioning" [10]. This algorithm changes the connectivity of a network and renders it singly connected. The resulting network is solved by Pearl's algorithm. The complexity of this method grows exponentially with the size of the loop cutset for a multiple connected network. Unfortunately the loop cutset minimization problem is NP-hard.

Another exact inference algorithm called "clique-tree propagation" [14] transforms a multiple connected network into a clique tree, then it performs message propagation method over the transformed network. The clique-tree propagation algorithm can be extremely slow for dense networks since its complexity is exponential with the size of the largest clique of the transformed network.

The approximate inference algorithm remains the best alternative of any exact inference algorithm when it is difficult or impossible to apply an exact inference method.

## 4. Loopy belief propagation

The "Loopy belief propagation" (LBP) is an approximate inference algorithm which applies the rules of belief propagation over networks with loops [7]. The main idea of LBP is to keep passing messages around the network until a stable belief state is reached (if ever). The LBP algorithm may not give exact results on a network with loops, but it can be used in the following way: iterate message propagation until convergence.

Empirically, several applications of the LBP algorithm proved successful, among them the decoding Turbo code: the error correction codes network [1].

We define here some notations used in the algorithm:

$\lambda_Y(X)$ is defined as the message to X from a child node Y.

$\pi_X(U)$ is defined as the message to X from its parent U.

$\lambda_X(X)$ is a message sent by X to itself if it is observed ($X \in E$).

We allow messages to change at each iteration t: $\lambda^{(t)}(X)$ is a message at iteration t.

Belief is the normalized product of all messages after convergence:

$BEL(x) = \alpha \lambda(x) \pi(x) \approx P(X=x|E)$.

To represent the messages propagation between the nodes, we draw the incoming messages exchanged at step t for a node X with parents $U=\{U_1,...U_n\}$ and children $Y=\{Y_1,...Y_m\}$ (see Figure 3):

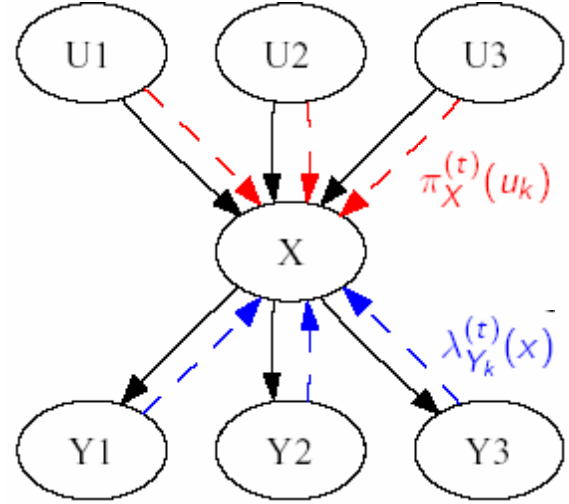

Figure 3: Incoming messages to node X at step t

with :
$$\lambda^{(t)}(x) = \lambda_X(x) \prod_j \lambda_{Y_j}^{(t)}(x) \qquad (1)$$

and :
$$\pi^{(t)}(x) = \sum_u P(X=x|U=u) \prod_k \pi_X^{(t)}(u_k) \qquad (2)$$

Similarly, we draw the outgoing messages exchanged at step t+1 from a node X to its parents and children (see Figure 4):

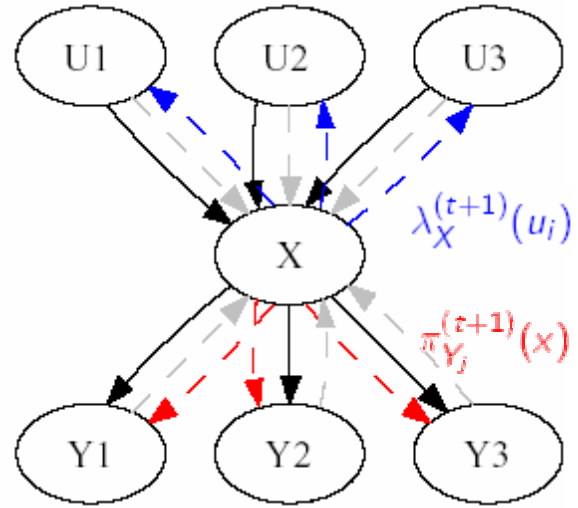

Figure 4 : Outgoing messages from node X at step t + 1

with :

$$\lambda_X^{(t+1)}(u_i) = \alpha \sum_x \lambda^{(t)}(x) \sum_{u_k : k \neq i} P(x|u) \prod_{k \neq i} \pi_X^{(t)}(u_k)$$

and :

$$\pi_{Y_j}^{(t+1)}(x) = \alpha \pi^{(t)}(x) \lambda_X(x) \prod_{k \neq j} \lambda_{Y_k}^{(t)}(x)$$

Nodes are updated in parallel: at each iteration, all nodes compute their outgoing messages based on the input of their neighbours from the previous iteration. The messages are said to converge if none of the beliefs in successive iterations changed by more than a small threshold (e.g. $10^{-4}$) [7].

When LBP algorithm converges, the provided posterior probabilities values are often a good approximation to the exact inference result. But if it does not converge it may oscillate between two belief states.

Murphy and al. tested LBP algorithm over both synthetic (PYRAMID and toyQMR), and real word (ALARM and QMR-DT) network [7]. They noted that LBP converges for all network except QMR-DT in which the algorithm oscillates between two belief states. They explain this result by the low prior probabilities values and the absence of randomization in the QMR-DT network.

They tried to avoid oscillations by using "momentum" term in equation 1 and 2. In general, they found that momentum significantly reduced the chance of oscillation. However, in several cases the beliefs to which the algorithm converged were quite inaccurate.

## 5. Introduction to the possibility theory

Possibility theory, introduced by Zadeh [16] and developed by Dubois and Prade [3], treats uncertainty in a qualitative or quantitative way. Uncertainty in possibility theory is represented by a pair of dual "measures" of possibility and necessity, usually graded on the unit interval called possibilistic scale. Possibility measures are max-decomposable for the union of events, in contrast with probability measures which are additive, while necessity measures are min-decomposable for the intersection of events.

The possibility theory, resulting from the fuzzy set theory, allows a better flexibility in the treatment of information available than within the probabilistic framework. It differs from the probability theory especially by the fact that it is possible to distinguish uncertainty from imprecise, which is not the case with probabilities [2].

We can distinguish between qualitative and quantitative possibility theories. Qualitative possibility theory can be defined in purely ordinal settings, while quantitative possibility theory requires the use of a numerical scale. Quantitative possibility measures can be viewed as upper bounds of imprecisely known probability measures. Several operational semantics for possibility degrees have been recently obtained. Qualitative and quantitative possibility theories differ in the way conditioning is defined (it is respectively based on minimum and product operations).

A logical counterpart of possibility theory has been developed for almost twenty years, and is known as possibilistic logic. A possibilistic logic formula is a pair of a classical logic formula and a weight understood as a lower bound of a necessity measure. Various extensions of possibilistic logic handle lower bounds of guaranteed or ordinary possibility functions, weights involving variables, fuzzy constants, multiple source information. Graphical representations of possibilistic logic bases, using the two types of conditioning, have been also obtained [2].

## 6. Conclusion and future work

In this paper we have presented two inference algorithms: the first, Pearl's belief propagation, is an exact algorithm applied to bayesian networks without loop. The second, LBP is an approximate inference algorithm which applies the rules of belief propagation over networks with loops. LBP algorithm guarantees a good approximation for posterior probabilities when it converges. However, convergence is not guaranteed when LBP is applied over QMR-DT network. We estimate that the transformation of the inference computation problem from the probability theory to the possibility theory will allow to improve LBP algorithm performances. We are Currently studying this transformation which we didn't try out yet.